\documentclass[twoside,11pt]{article}

\usepackage{jmlr2e}
\usepackage{color}
\usepackage{amsmath}
\usepackage{xspace}
\usepackage[dvipsnames]{xcolor}
\usepackage{subfigure}
\usepackage{multirow}
\usepackage{booktabs}

\usepackage[scaled]{beramono}
\usepackage{listings}
\usepackage{wrapfig}
\usepackage{capt-of}
\usepackage{caption}
\usepackage{fontawesome}
\usepackage{pifont}
\usepackage{array}
\usepackage{wasysym}

\hypersetup{citecolor=black, linkcolor=black}

\DeclareFixedFont{\ttb}{T1}{txtt}{bx}{n}{9}
\DeclareFixedFont{\ttm}{T1}{txtt}{m}{n}{9}

\definecolor{deepblue}{rgb}{0,0,0.5}
\definecolor{deepred}{rgb}{0.6,0,0}
\definecolor{deepgreen}{rgb}{0,0.5,0}

\lstdefinestyle{RStyle} {
  language=R,
  showstringspaces=false,
  formfeed=newpage,
  tabsize=4,
  commentstyle=\color{deepgreen},
  keywordstyle=\ttb\color{blue},
  stringstyle=\color{deepred},
  numberstyle=\scriptsize \color{black},
  basicstyle=\ttm,
  breakatwhitespace=false,
  breaklines=true,
  captionpos=b,
  keepspaces=true,
  numbers=left,
  numbersep=8pt,
  showspaces=false,          
  showstringspaces=false,
  showtabs=false,
  otherkeywords={!,!=,~,*,\&,\%/\%,\%*\%,\%\%,<<-,/,cat,library},
  alsoother={.,_,data,c,beta,family,ts,t,round},
}

\newcommand{\abess}{\textsf{abess}\xspace}

\newcommand{\elasticnet}{\textsf{elasticnet}\xspace}
\newcommand{\print}{\textsf{print}\xspace}
\newcommand{\coef}{\textsf{coef}\xspace}
\newcommand{\plot}{\textsf{plot}\xspace}
\newcommand{\tidyverse}{\textsf{tidyverse}\xspace}

\newcommand{\sklearn}{\textsf{scikit-learn}\xspace}

\newcommand{\PEP}{\textsf{PEP8}\xspace}
\newcommand{\sklearnbinomiallasso}{\textsf{LogisticRegressionCV}\xspace}
\newcommand{\sklearngaussianlasso}{\textsf{LassoCV}\xspace}
\newcommand{\celer}{\textsf{celer}\xspace}
\newcommand{\celergaussianlasso}{\textsf{LassoCV}\xspace}
\newcommand{\celerbinomiallasso}{\textsf{LogisticRegression}\xspace}
\newcommand{\sklearngaussianomp}{\textsf{OrthogonalMatchingPursuit}\xspace}

\newcommand{\ram}{\faStackOverflow}
\newcommand{\na}{{\ding{55}}}

\ShortHeadings{Abess---Fast Best-Subset Selection}{Zhu, Wang, Hu, Huang, Jiang, Zhang, Lin and Zhu}
\firstpageno{1}

\begin{document}

\title{abess: A Fast Best-Subset Selection Library in Python and R}
\author{\name Jin Zhu\textsuperscript{1} \email zhuj37@mail2.sysu.edu.cn \\
\name Xueqin Wang\textsuperscript{2} \email wangxq20@ustc.edu.cn \\
\name Liyuan Hu\textsuperscript{1} \email huly5@mail2.sysu.edu.cn     \\
\name Junhao Huang\textsuperscript{1} \email huangjh256@mail2.sysu.edu.cn     \\
\name Kangkang Jiang\textsuperscript{1} \email jiangkk3@mail2.sysu.edu.cn \\
\name Yanhang Zhang\textsuperscript{3} \email zhangyh98@ruc.edu.cn \\
\name Shiyun Lin\textsuperscript{4} \email shiyunlin@stu.pku.edu.cn     \\
\name Junxian Zhu\textsuperscript{5} \email junxian@nus.edu.sg \\
\textsuperscript{1} \addr Department of Statistical Science, Sun Yat-Sen University, Guangzhou, GD, China \\
\textsuperscript{2} \addr Department of Statistics and Finance/International Institute of Finance, School of Management, University of Science and Technology of China, Hefei, Anhui, China \\
\textsuperscript{3} \addr School of Statistics, Renmin University of China, Beijing, China \\
\textsuperscript{4} \addr Center for Statistical Science, Peking University, Beijing, China \\
\textsuperscript{5} \addr Saw Swee Hock School of Public Health, National University of Singapore, Singapore
}
\editor{}
\maketitle
\begingroup\renewcommand\thefootnote{*}
\footnotetext{Jin Zhu and Liyuan Hu contributed equally. Xueqin Wang is the corresponding author.}
\endgroup
\vspace{-10pt}
\begin{abstract}
  We introduce a new library named \abess that implements a unified framework of best-subset selection for solving diverse machine learning problems, e.g., linear regression, classification, and principal component analysis.
  Particularly, \abess certifiably gets the optimal solution within polynomial time with high probability under the linear model. Our efficient implementation allows \abess to attain the solution of best-subset selection problems as fast as or even \textsf{20x} faster than existing competing variable (model) selection toolboxes.
  Furthermore, it supports common variants like best subset of groups selection and $\ell_2$ regularized best-subset selection.  
  The core of the library is programmed in C++.
  For ease of use, a Python library is designed for convenient integration with \sklearn,
  and it can be installed from the Python Package Index (PyPI). In addition, a user-friendly
  R library is available at the Comprehensive R Archive Network (CRAN).
  The source code is available at: \url{https://github.com/abess-team/abess}.
\end{abstract}


\vspace{-5pt}
\section{Introduction}
\vspace{-2pt}
Best-subset selection (BSS) is imperative in machine learning and statistics.
It aims to find a minimally adequate subset of variables to accurately fit the data,
naturally reflecting the Occam's razor principle of simplicity.
Nowadays, the BSS also has far-reaching applications in every facet of research like medicine and biology because of the surge of large-scale datasets across a variety of fields.
As a benchmark optimization problem in machine learning and statistics, the BSS is also well-known as an NP-hard problem \citep{natarajan1995sparse}.
However, recent progress shows that the BSS can be efficiently solved \citep{huang2018constructive, zhu2020abess, prokopyev2021mifo}.
Especially, the ABESS algorithm using a splicing technique finds the best subset under the classical linear model in polynomial time with high probability \citep{zhu2020abess},
making itself even more attractive to practitioners.

We present a new library named \abess that implements a unified toolkit based on the splicing technique proposed by \citet{zhu2020abess}.
The supported solvers in \abess are summarized in Table~\ref{tab:support-solvers}.
Furthermore, our implementation improves computational efficiency by warm-start initialization, sparse matrix support, and OpenMP parallelism.
\abess can run on most Linux distributions, Windows 32 or 64-bit, and macOS with Python (version $\geq 3.6$) or R (version $\geq 3.1.0$),
and can be easily installed from PyPI\footnote{\url{https://pypi.org/project/abess/}} and CRAN\footnote{\url{https://cran.r-project.org/web/packages/abess}}.
\abess provides complete documentation\footnote{\url{https://abess.readthedocs.io} and \url{https://abess-team.github.io/abess/}},
where the API reference presents the syntax and the tutorial presents comprehensible examples for new users.
It relies on Github Actions\footnote{\url{https://github.com/abess-team/abess/actions}} for continuous integration.
The \PEP/\tidyverse style guide keeps the source Python/R code clean.
Code quality is assessed by standard code coverage metrics \citep{myers2011art}.
The coverages for the Python and the R packages at the time of writing are 97\% and 96\%, respectively.
The source code is distributed under the GPL-3 license.
\begin{table}[htbp]
  \vspace*{-8pt}
  \centering
  {\scriptsize
    \begin{tabular}{c|c|c}
      \toprule
      {\raggedleft Learning}                                                            & Target                                                & Solver                   (Reference)                                      \\
      \midrule
      \multirow{8}{*}{\parbox{1.5cm}{Supervised (model the variation of response~$y$)}} & $y \in \mathbb{R}$                                    & \textsf{LinearRegression}       \citep{zhu2020abess}                      \\
                                                                                        & $y \in \{0, 1\}$                                      & \textsf{LogisticRegression}     \citep{bestlogistic1989hosmer}            \\
                                                                                        & $y \in \{0, 1, 2, \ldots \}$                          & \textsf{PoissonRegression}      \citep{vincent2010glmulti}                \\
                                                                                        & $y \in (0, \infty)$                                   & \textsf{GammaRegression}        \citep{vincent2010glmulti}                \\
                                                                                        & $y \in \{\textup{type1}, \textup{type2}, \ldots \}$   & \textsf{MultinomialRegression}  \citep{krishnapuram2005sparsemultinomial} \\
                                                                                        & $y \in \{\textup{level1}, \textup{level2}, \ldots \}$ & \textsf{OrdinalRegression}      \citep{JSSv099i06}                        \\
                                                                                        & $y \in \mathbb{R} \times \{0, 1\}$                    & \textsf{CoxPHSurvivalAnalysis}  \citep{sebastian2020sksurv}               \\
                                                                                        & $y \in \mathbb{R}^d$                                  & \textsf{MultiTaskRegression}    \citep{zhang2017multitask}                \\
      \midrule
      \multirow{2}{*}{\raggedleft Unsupervised}                                         & Dimension reduction                                   & \textsf{SparsePCA}              \citep{d2008optimal}                      \\
                                                                                        & Matrix decomposition                                  & \textsf{RobustPCA}              \citep{cai2019accelerated}                \\
      \bottomrule
    \end{tabular}
  }
  \vspace*{-5pt}
  \caption{The supported best-subset selection solvers. \textsf{PCA}: principal component analysis.}\label{tab:support-solvers}
  \vspace*{-18pt}
\end{table}

\vspace{-5.8pt}
\section{Architecture}
\vspace{-2pt}
\begin{figure}[htbp]
  \vspace{-35pt}
  \begin{center}
    \includegraphics[width=0.96\textwidth]{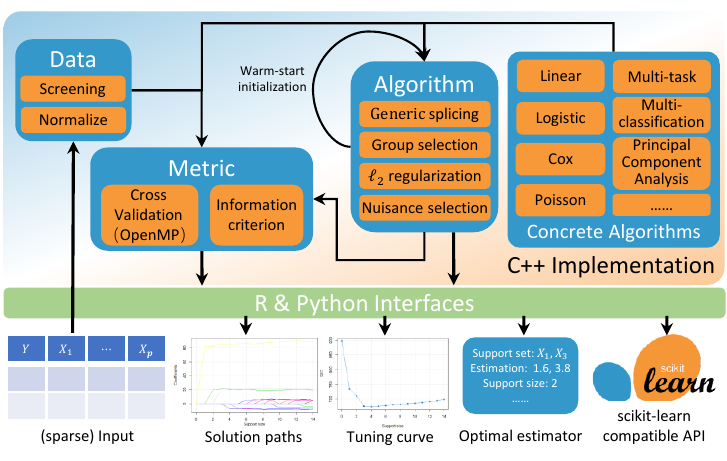}
  \end{center}
  \vspace*{-24pt}
  \caption{\abess software architecture.}\label{fig:abess-architecture}
  \vspace*{-15pt}
\end{figure}
Figure~\ref{fig:abess-architecture} shows the architecture of \abess, and each building block will be described as follows.
The \textbf{Data} class accepts the (sparse) tabular data from Python and R interfaces, and returns an object containing the predictors that are (optionally) screened \citep{fan2008sis} or normalized.
The \textbf{Algorithm} class implements the generic splicing technique for the BSS with the additional support for group-structure predictors \citep{zhang2021certifiably}, $\ell_2$-regularization for parameters \citep{dimitri2020sparse}, and nuisance selection \citep{sun2020nuisance}.
The concrete algorithms are programmed as subclasses of \textbf{Algorithm} by rewriting the virtual function interfaces of class \textbf{Algorithm}.
Seven implemented BSS tasks
are presented in Figure~\ref{fig:abess-architecture}.
Beyond that, the modularized design facilitates users to extend the library to various machine learning tasks by writing a subclass of \textbf{Algorithm}.
The \textbf{Metric} class assesses the estimation returned by the \textbf{Algorithm} class by the cross validation or information criteria like the Akaike information criterion and the high dimensional Bayesian information criterion \citep{akaike1998information, hbic2013wang}.
%
%
Python and R interfaces collect and process the results of the \textbf{Algorithm} and \textbf{Metric} classes.
The \abess Python library is compatible with \sklearn \citep{fabian2011sklearn}.
For each solver (e.g., \textsf{LinearRegression}) in \abess, Python users can not only use a familiar \sklearn API to train the model but also easily create a \sklearn pipeline including the model.
In the R library, S3 methods are programmed such that generic functions (like \print, \coef, and \plot) can be directly used to attain the BSS results and visualize solution paths or tuning value curves.

\vspace{-5.8pt}
\section{Usage Examples}
\vspace{-2pt}
Figure~\ref{fig:usage-example-optimality} shows that the \abess R library exactly selects
the effective variables and accurately estimates the coefficients.
Figure~\ref{fig:usage-example} illustrates the integration of the \abess Python interface with
\sklearn's modules to build a non-linear model for diagnosing malignant tumors. The output of the code reports the information of the polynomial features for the selected model among candidates, and its corresponding area under the curve (AUC), which is 0.966, indicating the selected model would have an admirable contribution in practice.
\begin{figure}[htbp]
  \begin{scriptsize}
    \vspace*{-0.36cm}
    \begin{lstlisting}[style=RStyle]
library(abess)
dat <- generate.data(n = 300, p = 1000, beta = c(3, -2, 0, 0, 2, rep(0, 995)))
best_est <- extract(abess(dat$x, dat$y, family = "gaussian"))
cat("Selected subset:", best_est$support.vars, 
    "and coefficient estimation:", round(best_est$support.beta, digits = 2))    
## Selected subset: x1 x2 x5 and coefficient estimation: 2.96 -2.05 1.9
\end{lstlisting}
    \vspace*{-0.5cm}
  \end{scriptsize}
  \caption{Using the \abess R library on a synthetic data set to demonstrate its optimality.
        The data set comes from a linear model with 
        the true sparse coefficients given by \textsf{beta}.
      }\label{fig:usage-example-optimality}
  \vspace*{-0.4cm}
\end{figure}
\begin{figure}[htbp]
  \begin{scriptsize}
    \vspace*{-1.2cm}
    \begin{lstlisting}[language=Python]
from abess.linear import LogisticRegression
from sklearn.datasets import load_breast_cancer
from sklearn.pipeline import Pipeline
from sklearn.metrics import make_scorer, roc_auc_score
from sklearn.preprocessing import PolynomialFeatures
from sklearn.model_selection import GridSearchCV
# combine feature transform and model:
pipe = Pipeline([('poly', PolynomialFeatures(include_bias=False)), 
  ('logreg', LogisticRegression())])
param_grid = {'poly__interaction_only':[True, False], 'poly__degree':[1, 2, 3]}
# Use cross validation to tune parameters:
scorer = make_scorer(roc_auc_score, greater_is_better=True)
grid_search = GridSearchCV(pipe, param_grid, scoring=scorer, cv=5)
# load and fitting example data set:
X, y = load_breast_cancer(return_X_y=True)
grid_search.fit(X, y)
# print the best tuning parameter and associated AUC score:
print([grid_search.best_params_, grid_search.best_score_])
# >>> [{'poly__degree': 2, 'poly__interaction_only': True}, 0.9663829492654472]
\end{lstlisting}
    \vspace*{-0.5cm}
  \end{scriptsize}
  \caption{Example of using the \abess Python library with \sklearn.}\label{fig:usage-example}
  \vspace*{-0.4cm}
\end{figure}







\vspace{-5.8pt}
\section{Performance}\label{sec:performance}
\vspace{-2pt}
We compare \abess with popular variable selection libraries in Python and R through regression, classification, and PCA.
The libraries include: \sklearn (a benchmark Python library for machine learning),
\celer (a fast Python solver for $\ell_1$-regularization optimization, \citet{pmlr-v80-massias18a, massias2020dual}),
and \elasticnet (a elastic-net R solver for sparse PCA, \citet{spca2006zou}).
All computations are conducted on a Ubuntu platform with Intel(R) Core(TM) i9-9940X CPU @ 3.30GHz and 48 RAM. Python version is 3.9.1 and R version is 3.6.3.
Table~\ref{tab:realdata-glm} displays the regression and classification analyse results,
suggesting \abess derives parsimonious models that achieve competitive performance in few minutes.
Particularly, for the cancer data set, it is more than \textsf{20x} faster than \sklearn ($\ell_1$).
The results of the sparse PCA (SPCA) are demonstrated in Table~\ref{tab:realdata-spca}.
Compared with \elasticnet, \abess consumes less than a tenth of its runtime but explains more variance under the same sparsity level.
\begin{table}[htbp]
  \vspace{-10pt}
    \begin{center}
      \begin{scriptsize}
        \begin{tabular}{c|c|ccc|ccc|ccc}
          \toprule
          \multirow{3}{*}{Library} & \multirow{3}{*}{\parbox{1cm}{\centering Version}}
                                   & \multicolumn{3}{c|}{Superconductivity}                   & \multicolumn{3}{c|}{Cancer}         & \multicolumn{3}{c}{Musk}                                                                                                               \\
                                   &                                                          & \multicolumn{3}{c|}{$(3895 \times 85400)$} & \multicolumn{3}{c|}{$(118 \times 22215)$} & \multicolumn{3}{c}{$(7074 \times 166)$}                                                    \\
                                   &                                                          & MSE                                        & NNZ                                       & Runtime                                 & AUC  & NNZ   & Runtime & AUC  & NNZ    & Runtime \\
          \midrule
          \sklearn ($\ell_1$)      & 1.0.0                                                    & 33.56                                      & 1126.70                                   & 1043.96                                 & 0.92 & 366.55 & 62.16   & 0.97 & 165.40 & 53.63 \\
          \celer                   & 0.6.1                                                    & 88.58                                      & 30.00                                     & 173.25                                  & 0.91 & 20.65 & 26.24  & 0.97 & 162.15 & 2.00 \\
          \sklearn ($\ell_0$)      & 1.0.0                                                    & \ram                                       & \ram                                      & \ram                                    & \na  & \na   & \na     & \na  & \na    & \na     \\
          \abess                   & 0.4.5                                                    & 41.72                                      & 81.50                                     & 110.41                                   & 0.96 & 1.00  & 2.91    & 0.97 & 155.25 & 140.68  \\
          \bottomrule
        \end{tabular}
      \end{scriptsize}
    \end{center}
  \vspace*{-20pt}
  \caption{Average performance on the superconductivity data set (for regression),
    the cancer and the musk data sets (for classification) \citep{chin2006cancer, dua2019uci, hamidieh2018data} based on 20 randomly drawn test sets.
    NNZ: the number of non-zero elements. 
    Runtime is measured in seconds.
      \sklearn ($\ell_1$): \sklearngaussianlasso (for regression) and  \sklearnbinomiallasso (for classification).
        \celer: \celergaussianlasso (for regression) and \celerbinomiallasso (for classification).
    \sklearn ($\ell_0$): \sklearngaussianomp (for regression).
    \na: not available. \ram: memory overflow. 
  }\label{tab:realdata-glm}
\end{table}
\begin{table}[htbp]
  \vspace{-22pt}
    \begin{center}
      \begin{scriptsize}
        \begin{tabular}{c|cc|cc|cc}
          \toprule
          Sparsity           & \multicolumn{2}{c|}{5} & \multicolumn{2}{c|}{10} & \multicolumn{2}{c}{20}                                 \\
          \hline
          Library            & \elasticnet            & \abess                  & \elasticnet            & \abess & \elasticnet & \abess \\
          \midrule
          Explained variance & 1.37                   & 2.28                    & 1.61                   & 3.03   & 2.00        & 3.88   \\
          Runtime (seconds)  & 15.87                  & 1.06                    & 15.77                  & 1.07   & 85.54       & 1.05   \\
          \bottomrule
        \end{tabular}
      \end{scriptsize}
    \end{center}
    \vspace*{-20pt}
    \caption{Performance of the SPCA when 5, 10, 20 elements in the loading vector of the first principal component are non-zero.
      The data set has 217 observations, where each observation has 1,413 genetic factors \citep{Christensen2009gu}.
      \elasticnet: version 1.3.0.
    }\label{tab:realdata-spca}
  \vspace*{-23pt}
\end{table}

\vspace{-3.8pt}
\section{Conclusion}
\vspace{-5pt}
\abess is a fast and comprehensive library for solving various BSS problems with statistical guarantees. It offers user-friendly interfaces for both Python and R users, and seamlessly integrates with existing ecosystems.
Therefore, the \abess library is a potentially indispensable toolbox for machine learning and related applications.
Future versions of \abess intend to support other important machine learning tasks, and adapt to advanced machine learning pipelines in Python and R \citep{mlr3, feurer2021openml, mlr3pipelines}.

\acks{We would like to thank three reviewers for their constructive suggestions and valuable comments, which have
  substantially improved this article and the \abess library.
  Wang's research is partially supported by NSFC (72171216, 71921001, 71991474), 
  The Key Research and Development Program of Guangdong, China (2019B020228001), and
  Science and Technology Program of Guangzhou, China (202002030129).
  Zhu's research is partially supported by the Outstanding Graduate Student Innovation and Development Program of Sun Yat-Sen University (19lgyjs64).
  Zhang's research is supported by the Fundamental Research Funds for the Central Universities, and the Research Funds of Renmin University of China (22XNH161).
  We are grateful to UCI Machine Learning Repository for sharing the superconductivity and musk data sets.
}










\hypersetup{colorlinks=true, urlcolor=black, citecolor=black, linkcolor=black}
\vskip 0.2in
\bibliography{reference}

\begin{thebibliography}{30}
\providecommand{\natexlab}[1]{#1}
\providecommand{\url}[1]{\texttt{#1}}
\expandafter\ifx\csname urlstyle\endcsname\relax
  \providecommand{\doi}[1]{doi: #1}\else
  \providecommand{\doi}{doi: \begingroup \urlstyle{rm}\Url}\fi

\bibitem[Akaike(1998)]{akaike1998information}
Hirotogu Akaike.
\newblock Information theory and an extension of the maximum likelihood
  principle.
\newblock In \emph{Selected Papers of Hirotugu Akaike}, pages 199--213.
  Springer, 1998.

\bibitem[Bertsimas and Parys(2020)]{dimitri2020sparse}
Dimitris Bertsimas and Bart~Van Parys.
\newblock Sparse high-dimensional regression: Exact scalable algorithms and
  phase transitions.
\newblock \emph{The Annals of Statistics}, 48\penalty0 (1):\penalty0 300 --
  323, 2020.
\newblock \doi{10.1214/18-AOS1804}.
\newblock URL \url{https://doi.org/10.1214/18-AOS1804}.

\bibitem[Binder et~al.(2021)Binder, Pfisterer, Lang, Schneider, Kotthoff, and
  Bischl]{mlr3pipelines}
Martin Binder, Florian Pfisterer, Michel Lang, Lennart Schneider, Lars
  Kotthoff, and Bernd Bischl.
\newblock {mlr3pipelines} -- flexible machine learning pipelines in r.
\newblock \emph{Journal of Machine Learning Research}, 22\penalty0
  (184):\penalty0 1--7, 2021.
\newblock URL \url{http://jmlr.org/papers/v22/21-0281.html}.

\bibitem[Cai et~al.(2019)Cai, Cai, and Wei]{cai2019accelerated}
Hanqin Cai, Jianfeng Cai, and Ke~Wei.
\newblock Accelerated alternating projections for robust principal component
  analysis.
\newblock \emph{Journal of Machine Learning Research}, 20\penalty0
  (1):\penalty0 685--717, 2019.

\bibitem[Chin et~al.(2006)Chin, DeVries, Fridlyand, Spellman, Roydasgupta, Kuo,
  Lapuk, Neve, Qian, Ryder, Chen, Feiler, Tokuyasu, Kingsley, Dairkee, Meng,
  Chew, Pinkel, Jain, Ljung, Esserman, Albertson, Waldman, and
  Gray]{chin2006cancer}
Koei Chin, Sandy DeVries, Jane Fridlyand, Paul~T Spellman, Ritu Roydasgupta,
  Wen-Lin Kuo, Anna Lapuk, Richard~M Neve, Zuwei Qian, Tom Ryder, Fanqing Chen,
  Heidi Feiler, Taku Tokuyasu, Chris Kingsley, Shanaz Dairkee, Zhenhang Meng,
  Karen Chew, Daniel Pinkel, Ajay Jain, Britt~Marie Ljung, Laura Esserman,
  Donna~G Albertson, Frederic~M Waldman, and Joe~W Gray.
\newblock {Genomic and transcriptional aberrations linked to breast cancer
  pathophysiologies}.
\newblock \emph{Cancer Cell}, 10\penalty0 (6):\penalty0 529--541, December
  2006.

\bibitem[Christensen et~al.(2009)Christensen, Houseman, Marsit, Zheng, Wrensch,
  Wiemels, Nelson, Karagas, Padbury, Bueno, Sugarbaker, Yeh, Wiencke, and
  Kelsey]{Christensen2009gu}
Brock~C Christensen, E~Andres Houseman, Carmen~J Marsit, Shichun Zheng,
  Margaret~R Wrensch, Joseph~L Wiemels, Heather~H Nelson, Margaret~R Karagas,
  James~F Padbury, Raphael Bueno, David~J Sugarbaker, Ru-Fang Yeh, John~K
  Wiencke, and Karl~T Kelsey.
\newblock {Aging and Environmental Exposures Alter Tissue-Specific DNA
  Methylation Dependent upon CpG Island Context}.
\newblock \emph{PLOS Genetics}, 5\penalty0 (8):\penalty0 e1000602, August 2009.

\bibitem[d'Aspremont et~al.(2008)d'Aspremont, Bach, and
  El~Ghaoui]{d2008optimal}
Alexandre d'Aspremont, Francis Bach, and Laurent El~Ghaoui.
\newblock Optimal solutions for sparse principal component analysis.
\newblock \emph{Journal of Machine Learning Research}, 9\penalty0 (7), 2008.

\bibitem[Dua and Graff(2017)]{dua2019uci}
Dheeru Dua and Casey Graff.
\newblock {UCI} machine learning repository, 2017.
\newblock URL \url{http://archive.ics.uci.edu/ml}.

\bibitem[Fan and Lv(2008)]{fan2008sis}
Jianqing Fan and Jinchi Lv.
\newblock Sure independence screening for ultrahigh dimensional feature space.
\newblock \emph{Journal of the Royal Statistical Society: Series B (Statistical
  Methodology)}, 70\penalty0 (5):\penalty0 849--911, 2008.
\newblock \doi{https://doi.org/10.1111/j.1467-9868.2008.00674.x}.

\bibitem[Feurer et~al.(2021)Feurer, van Rijn, Kadra, Gijsbers, Mallik, Ravi,
  Müller, Vanschoren, and Hutter]{feurer2021openml}
Matthias Feurer, Jan~N. van Rijn, Arlind Kadra, Pieter Gijsbers, Neeratyoy
  Mallik, Sahithya Ravi, Andreas Müller, Joaquin Vanschoren, and Frank Hutter.
\newblock Openml-python: an extensible python api for openml.
\newblock \emph{Journal of Machine Learning Research}, 22\penalty0
  (100):\penalty0 1--5, 2021.
\newblock URL \url{http://jmlr.org/papers/v22/19-920.html}.

\bibitem[G{\'o}mez and Prokopyev(2021)]{prokopyev2021mifo}
Andr{\'e}s G{\'o}mez and Oleg~A. Prokopyev.
\newblock A mixed-integer fractional optimization approach to best subset
  selection.
\newblock \emph{INFORMS Journal on Computing}, 33\penalty0 (2):\penalty0
  551--565, 2021.
\newblock \doi{10.1287/ijoc.2020.1031}.

\bibitem[Hamidieh(2018)]{hamidieh2018data}
Kam Hamidieh.
\newblock A data-driven statistical model for predicting the critical
  temperature of a superconductor.
\newblock \emph{Computational Materials Science}, 154:\penalty0 346--354, 2018.

\bibitem[Hosmer et~al.(1989)Hosmer, Jovanovic, and
  Lemeshow]{bestlogistic1989hosmer}
David~W. Hosmer, Borko Jovanovic, and Stanley Lemeshow.
\newblock Best subsets logistic regression.
\newblock \emph{Biometrics}, 45\penalty0 (4):\penalty0 1265--1270, 1989.
\newblock URL \url{http://www.jstor.org/stable/2531779}.

\bibitem[Huang et~al.(2018)Huang, Jiao, Liu, and Lu]{huang2018constructive}
Jian Huang, Yuling Jiao, Yanyan Liu, and Xiliang Lu.
\newblock A constructive approach to $l_0$ penalized regression.
\newblock \emph{Journal of Machine Learning Research}, 19\penalty0
  (1):\penalty0 403--439, 2018.

\bibitem[Krishnapuram et~al.(2005)Krishnapuram, Carin, Figueiredo, and
  Hartemink]{krishnapuram2005sparsemultinomial}
Balaji Krishnapuram, Lawrence Carin, Mario A.~T. Figueiredo, and Alexander~J.
  Hartemink.
\newblock Sparse multinomial logistic regression: fast algorithms and
  generalization bounds.
\newblock \emph{IEEE Transactions on Pattern Analysis and Machine
  Intelligence}, 27\penalty0 (6):\penalty0 957--968, 2005.
\newblock \doi{10.1109/TPAMI.2005.127}.

\bibitem[Lang et~al.(2019)Lang, Binder, Richter, Schratz, Pfisterer, Coors, Au,
  Casalicchio, Kotthoff, and Bischl]{mlr3}
Michel Lang, Martin Binder, Jakob Richter, Patrick Schratz, Florian Pfisterer,
  Stefan Coors, Quay Au, Giuseppe Casalicchio, Lars Kotthoff, and Bernd Bischl.
\newblock {mlr3}: A modern object-oriented machine learning framework in {R}.
\newblock \emph{Journal of Open Source Software}, dec 2019.
\newblock \doi{10.21105/joss.01903}.

\bibitem[Massias et~al.(2018)Massias, Gramfort, and
  Salmon]{pmlr-v80-massias18a}
Mathurin Massias, Alexandre Gramfort, and Joseph Salmon.
\newblock Celer: a fast solver for the lasso with dual extrapolation.
\newblock In \emph{International Conference on Machine Learning}, volume~80,
  pages 3321--3330, 2018.

\bibitem[Massias et~al.(2020)Massias, Vaiter, Gramfort, and
  Salmon]{massias2020dual}
Mathurin Massias, Samuel Vaiter, Alexandre Gramfort, and Joseph Salmon.
\newblock Dual extrapolation for sparse glms.
\newblock \emph{Journal of Machine Learning Research}, 21\penalty0
  (234):\penalty0 1--33, 2020.
\newblock URL \url{http://jmlr.org/papers/v21/19-587.html}.

\bibitem[Myers et~al.(2011)Myers, Sandler, and Badgett]{myers2011art}
Glenford~J Myers, Corey Sandler, and Tom Badgett.
\newblock \emph{The Art of Software Testing}.
\newblock John Wiley \& Sons, 2011.

\bibitem[Natarajan(1995)]{natarajan1995sparse}
Balas~Kausik Natarajan.
\newblock Sparse approximate solutions to linear systems.
\newblock \emph{SIAM Journal on Computing}, 24\penalty0 (2):\penalty0 227--234,
  1995.
\newblock \doi{10.1137/S0097539792240406}.

\bibitem[Pedregosa et~al.(2011)Pedregosa, Varoquaux, Gramfort, Michel, Thirion,
  Grisel, Blondel, Prettenhofer, Weiss, Dubourg, Vanderplas, Passos,
  Cournapeau, Brucher, Perrot, and {{\'E}}douard Duchesnay]{fabian2011sklearn}
Fabian Pedregosa, Ga{{\"e}}l Varoquaux, Alexandre Gramfort, Vincent Michel,
  Bertrand Thirion, Olivier Grisel, Mathieu Blondel, Peter Prettenhofer, Ron
  Weiss, Vincent Dubourg, Jake Vanderplas, Alexandre Passos, David Cournapeau,
  Matthieu Brucher, Matthieu Perrot, and {{\'E}}douard Duchesnay.
\newblock Scikit-learn: Machine learning in python.
\newblock \emph{Journal of Machine Learning Research}, 12\penalty0
  (85):\penalty0 2825--2830, 2011.
\newblock URL \url{http://jmlr.org/papers/v12/pedregosa11a.html}.

\bibitem[P{\"o}lsterl(2020)]{sebastian2020sksurv}
Sebastian P{\"o}lsterl.
\newblock scikit-survival: A library for time-to-event analysis built on top of
  scikit-learn.
\newblock \emph{Journal of Machine Learning Research}, 21\penalty0
  (212):\penalty0 1--6, 2020.
\newblock URL \url{http://jmlr.org/papers/v21/20-729.html}.

\bibitem[Sun and Zhang(2021)]{sun2020nuisance}
Qiang Sun and Heping Zhang.
\newblock Targeted inference involving high-dimensional data using nuisance
  penalized regression.
\newblock \emph{Journal of the American Statistical Association}, 116\penalty0
  (535):\penalty0 1472--1486, 2021.
\newblock \doi{10.1080/01621459.2020.1737079}.

\bibitem[Vincent and Claire(2010)]{vincent2010glmulti}
Calcagno Vincent and de~Mazancourt Claire.
\newblock glmulti: An r package for easy automated model selection with
  (generalized) linear models.
\newblock \emph{Journal of Statistical Software}, 34\penalty0 (12):\penalty0
  1--29, 2010.
\newblock \doi{10.18637/jss.v034.i12}.

\bibitem[Wang et~al.(2013)Wang, Kim, and Li]{hbic2013wang}
Lan Wang, Yongdai Kim, and Runze Li.
\newblock {Calibrating nonconvex penalized regression in ultra-high dimension}.
\newblock \emph{The Annals of Statistics}, 41\penalty0 (5):\penalty0 2505 --
  2536, 2013.
\newblock \doi{10.1214/13-AOS1159}.

\bibitem[Wurm et~al.(2021)Wurm, Rathouz, and Hanlon]{JSSv099i06}
Michael~J. Wurm, Paul~J. Rathouz, and Bret~M. Hanlon.
\newblock Regularized ordinal regression and the ordinalnet r package.
\newblock \emph{Journal of Statistical Software}, 99\penalty0 (6):\penalty0
  1--42, 2021.
\newblock \doi{10.18637/jss.v099.i06}.

\bibitem[Zhang et~al.(2021)Zhang, Zhu, Zhu, and Wang]{zhang2021certifiably}
Yanhang Zhang, Junxian Zhu, Jin Zhu, and Xueqin Wang.
\newblock A splicing approach to best subset of groups selection.
\newblock \emph{arXiv preprint arXiv:2104.12576}, 2021.

\bibitem[Zhang and Yang(2017)]{zhang2017multitask}
Yu~Zhang and Qiang Yang.
\newblock An overview of multi-task learning.
\newblock \emph{National Science Review}, 5\penalty0 (1):\penalty0 30--43, 09
  2017.
\newblock ISSN 2095-5138.
\newblock \doi{10.1093/nsr/nwx105}.

\bibitem[Zhu et~al.(2020)Zhu, Wen, Zhu, Zhang, and Wang]{zhu2020abess}
Junxian Zhu, Canhong Wen, Jin Zhu, Heping Zhang, and Xueqin Wang.
\newblock A polynomial algorithm for best-subset selection problem.
\newblock \emph{Proceedings of the National Academy of Sciences}, 2020.
\newblock \doi{10.1073/pnas.2014241117}.
\newblock URL \url{https://www.pnas.org/doi/abs/10.1073/pnas.2014241117}.

\bibitem[Zou et~al.(2006)Zou, Hastie, and Tibshirani]{spca2006zou}
Hui Zou, Trevor Hastie, and Robert Tibshirani.
\newblock Sparse principal component analysis.
\newblock \emph{Journal of Computational and Graphical Statistics}, 15\penalty0
  (2):\penalty0 265--286, 2006.
\newblock \doi{10.1198/106186006X113430}.
\newblock URL \url{https://doi.org/10.1198/106186006X113430}.

\end{thebibliography}

\end{document}